\documentclass[letterpaper]{article} % DO NOT CHANGE THIS
\usepackage{aaai24}  % DO NOT CHANGE THIS
\usepackage{times}  % DO NOT CHANGE THIS
\usepackage{helvet}  % DO NOT CHANGE THIS
\usepackage{courier}  % DO NOT CHANGE THIS
\usepackage[hyphens]{url}  % DO NOT CHANGE THIS
\usepackage{graphicx} % DO NOT CHANGE THIS
\usepackage{natbib}  % DO NOT CHANGE THIS AND DO NOT ADD ANY OPTIONS TO IT
\usepackage{caption} % DO NOT CHANGE THIS AND DO NOT ADD ANY OPTIONS TO IT
\usepackage{algorithm}
\usepackage{algorithmic}

\usepackage{newfloat}
\usepackage{listings}
\DeclareCaptionStyle{ruled}{labelfont=normalfont,labelsep=colon,strut=off} % DO NOT CHANGE THIS
\floatstyle{ruled}
\newfloat{listing}{tb}{lst}{}
\floatname{listing}{Listing}
\usepackage{amssymb}
\usepackage{booktabs}
\usepackage{multirow}
\usepackage{amsmath}
\usepackage{makecell}

\title{Fine-Grained Multi-View Hand Reconstruction Using Inverse Rendering}
\author{
    Qijun Gan,
    Wentong Li,
    Jinwei Ren,
    Jianke Zhu\thanks{Corresponding author.} 
}
\affiliations{
    College of Computer Science and Technology, Zhejiang University, China\\
    \{ganqijun,liwentong,zijinxuxu,jkzhu\}@zju.edu.cn
}

\usepackage{bibentry}
\begin{document}

\maketitle

\begin{abstract}
Reconstructing high-fidelity hand models with intricate textures plays a crucial role in enhancing human-object interaction and advancing real-world applications. Despite the state-of-the-art methods excelling in texture generation and image rendering, they often face challenges in accurately capturing geometric details. Learning-based approaches usually offer better robustness and faster inference, which tend to produce smoother results and require substantial amounts of training data. To address these issues, we present a novel fine-grained multi-view hand mesh reconstruction method that leverages inverse rendering to restore hand poses and intricate details. Firstly, our approach predicts a parametric hand mesh model through Graph Convolutional Networks (GCN) based method from multi-view images. We further introduce a novel Hand Albedo and Mesh (HAM) optimization module to refine both the hand mesh and textures, which is capable of preserving the mesh topology. In addition, we suggest an effective mesh-based neural rendering scheme to simultaneously generate photo-realistic image and optimize mesh geometry by fusing the pre-trained rendering network with vertex features. We conduct the comprehensive experiments on InterHand2.6M, DeepHandMesh and dataset collected by ourself, whose promising results show that our proposed approach outperforms the state-of-the-art methods on both reconstruction accuracy and rendering quality. Code and dataset are publicly available at \url{https://github.com/agnJason/FMHR}. 

\end{abstract}

%%%%%%%%% BODY TEXT
\section{Introduction}
\label{sec:intro}

3D human reconstruction has attracted considerable research attentions~\cite{saito2019pifu,peng2021animatable,bib:HumanNeRF,chen2021snarf,2021narf,bhatnagar2020loopreg} in recent years. While there have been promising advancements in reconstructing the human body and face~\cite{lei2023hierarchical, grassal2021neural, peng2021neural,xiu2022icon}, it still remains a formidable challenge to achieve highly accurate hand reconstruction due to the inherent complexity of joint variations. By taking consideration of the distinctive nature for hands, it is essential to investigate the hand geometry and rendering for obtaining the realistic and fine-grained representations.

Conventional model-based methods, such as MANO~\cite{bib:MANO} and Nimble~\cite{li2022nimble}, often rely on smoothing meshes and texture maps for hand representation. Nevertheless, it typically requires costly scanning data and artistic expertise in order to achieve intricate and personalized hand meshes with texture maps. Moreover, the complex nature of hand movements and challenges posed by occlusions hinder the faithful restoration of hand mesh. Model-free approaches like LISA~\cite{bib:LISA} aim to address these challenges by reconstructing coherent hands from image sequences, while HandAvatar~\cite{bib:handavatar} focuses on reconstructing and rendering hands in arbitrary poses by disentangling reflectance and lighting. Nonetheless, the resulting mesh often exhibits smoothness due to large pose variations across different frames. % To tackle self-occlusion-induced shadows, HARP~\cite{karunratanakul2022harp} presents an approach that restores both the albedo and normal map. While this effectively addresses the shadow issue, the texture mainly manifests within the normal map, resulting in a lack of fine details in the mesh geometry.

\begin{figure}
	\centering
        \includegraphics[width=0.415\textwidth]{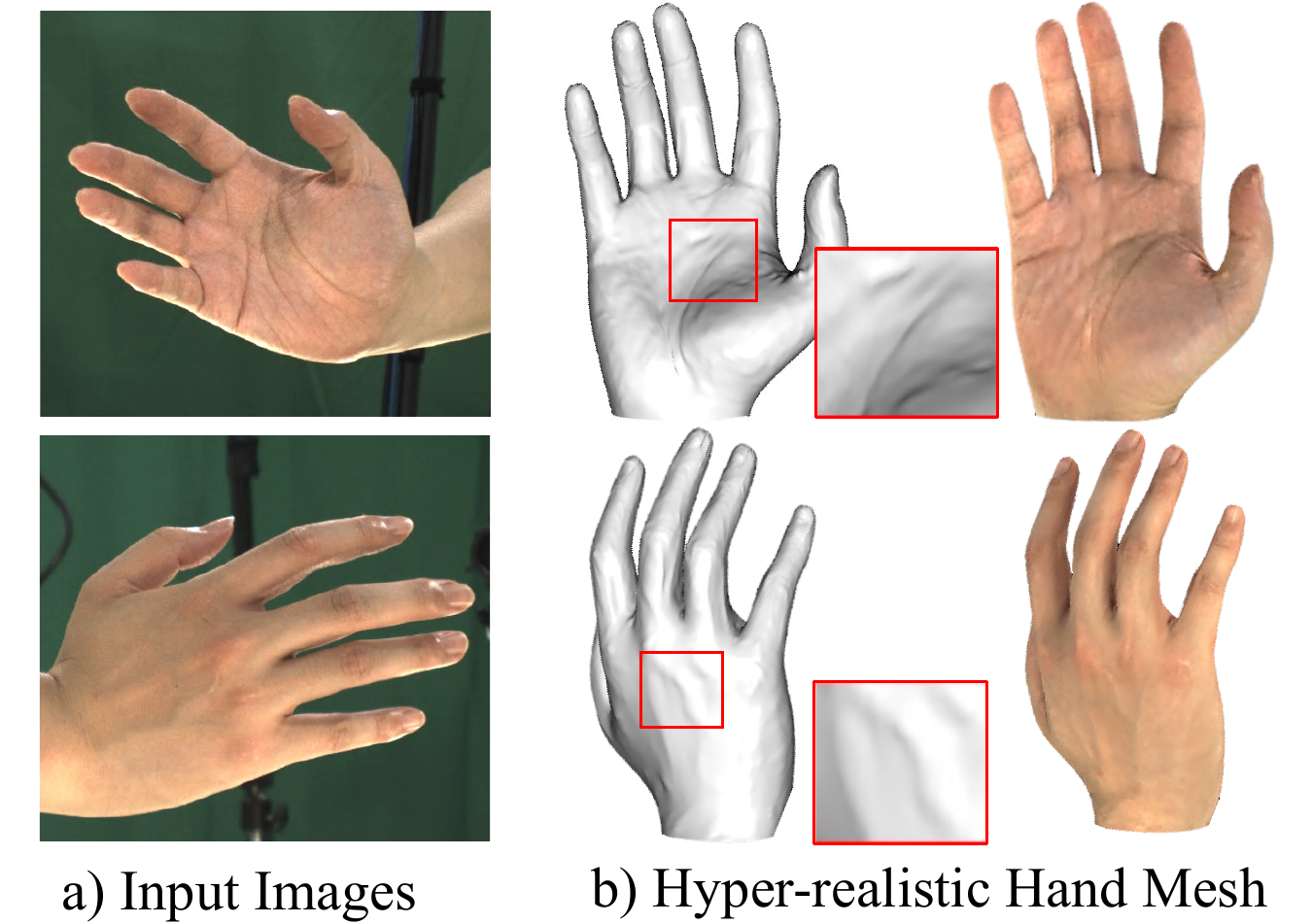}
        %\vspace{-0.05in}
	\caption{Our proposed approach focuses on reconstructing hands from multi-view images, allowing for the generation of precise poses, geometry, and photo-realistic rendering.}
	\label{fig:overall}
	%\vspace{-0.3in}
\end{figure}

Neural rendering-based methods, such as NeRF~\cite{NeRF} and NeuS~\cite{wang2021neus, Fu2022GeoNeus}, have been widely used in synthesizing static objects from multi-view images. As for the dynamic objects like hands, these methods have the difficulties in obtaining a fixed topological structure through implicit surface representation. Despite their impressive rendering results, NeuS~\cite{wang2021neus} requires a large amount of training time due to sampling along the ray. Meanwhile, NeuralBody~\cite{peng2021neural} attaches latent codes to SMPL~\cite{bib:SMPL} vertices, which enables to diffuse into space through sparse convolution. However, this may lead to some artifacts on the predicted mesh. To address these limitations, HandNeRF~\cite{handnerf} proposes a pose-driven deformation field to render photo-realistic hands from various views and poses, while it does not explicitly provide a hand mesh for animation and real-time rendering.

%{\bf UNCLEAR Additionally, the neural rendering based on static objects lacks hand structure representation, making it challenging for data driving.} H

%There has been a growing interest in neural rendering techniques, particularly in generating photorealistic renderings. Neural radiance fields~\cite{NeRF, Fu2022GeoNeus} have been utilized to represent humans by accurately predicting geometry and texture properties for any given 3D point query. However, these approaches face challenges in terms of hand topology and the generation of novel hand poses. HandNeRF~\cite{handnerf} successfully addresses these limitations and showcases realistic outcomes in hand rendering from various views and poses. Nevertheless, it does not offer a detailed representation of the hand in the form of mesh structure. NeuS~\cite{wang2021neus} showcases remarkable rendering capabilities; however, due to the constraints of neural rendering, generating individual models is time-consuming. Additionally, the neural rendering based on static objects lacks hand structure representation, making it challenging for data driving.   % story

To overcome the above challenges, we propose an effective coarse-to-fine approach to hand mesh reconstruction from multi-view images. By synergizing the benefits of parametric models and mesh-based rendering, our method achieves high-fidelity reconstruction results while maintaining fast training time with only 1.5 minutes. By incorporating multi-layer perceptrons into Graph Convolutional Networks (GCN)~\cite{gcn}, we can simultaneously recover hand mesh and MANO parameters. Unlike learning-based methods that demand extensive training data, we introduce a Hand Albedo and Mesh (HAM) optimization module by leveraging inverse rendering to enhance level of details. The resulting fine-grained mesh preserves the flexibility provided by the parametric model MANO~\cite{bib:MANO}, which enables to repose it into novel configurations. To further enhance rendering quality and refine the mesh, we suggest a mesh-based neural rendering scheme by fusing a pre-trained rendering network with vertex features. Our proposed method is evaluated on the InterHand2.6M, DeepHandMesh and dataset collected by ourself. The promising experimental results demonstrate its efficacy in reconstructing high-fidelity hands from multi-view images, as illustrated in Fig.~\ref{fig:overall}.% Our method outperforms other approaches, achieving higher quality in the reconstructed hand shapes.

%In summary, our proposed method represents a significant advancement in hand mesh reconstruction from multi-view images,  

Our main contributions are summarized as below.

\begin{itemize}
    \item We propose a coarse-to-fine approach to accurately recover the fine-grained hand mesh model from multi-view images by taking advantage of inverse rendering. %, we effectively capture personalized texture details.
    \item A novel HAM optimization module is presented to refine the over-smoothing results of parametric hand models.
    
    %To overcome the issue of excessive smoothness in the results, we combine the advantages of parametric hand models and our proposed HAM optimization module to achieve high-fidelity and fine-grained details in the reconstructed hand meshes.
    \item We devise an effective mesh-based neural rendering scheme to simultaneously generate photo-realistic image and optimize mesh geometry by fusing the pre-trained rendering network with vertex features.
    
    %Despite the considerable time demanded by conventional neural rendering methods, we develop a pre-trained mesh-based neural rendering approach that enables fast and efficient rendering, producing hyper-realistic results that surpass the current state-of-the-art methods.
    % \item Leveraging the benefits of the parametric model, our approach allows for easy reposing of the fine-detailed hand meshes to any novel pose.
\end{itemize}

%-------------------------------------------------------------------------
\section{Related Work}
\label{sec:relate}

\begin{figure*}
    \centering
    \includegraphics[width=0.92 \linewidth]{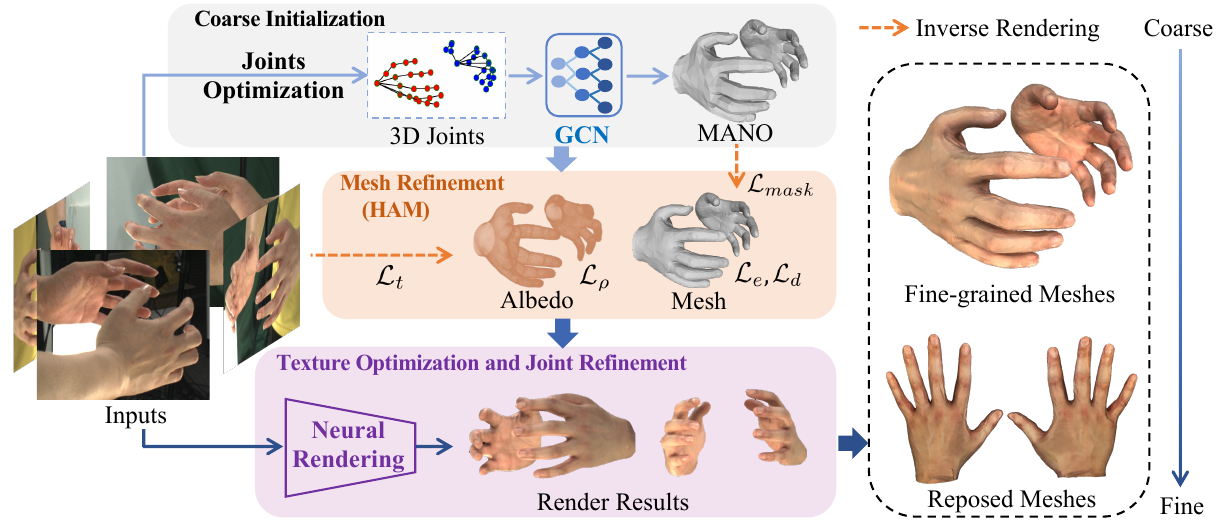}
    %\vspace{-0.2in}
    \caption{Overview of our coarse-to-fine framework. Given a set of calibrated images, we initialize MANO parameters and refine the mesh using our proposed HAM module and inverse rendering to achieve geometric details. By jointly optimizing the mesh using a model-based neural rendering, a fine-grained mesh can be obtained along with its hyper-realistic rendered images.} %{\bf Describe the method flow rather than discussing the results. } We observe hand poses from multi-views to generate precise MANO models with GCN. The hand's wrinkle details are restored using HAM. Our mesh-based neural rendering approach enables the creation of hyper-realistic hand renderings in new poses.} % no tech, need to change, 
    \label{fig:overview}
    %\vspace{-0.15in}
\end{figure*}

\subsection{Model-Based Hand Reconstruction}
Parametric models have been widely used in representing objects with the fixed typologies, such as the human body~\cite{bib:SMPL,bib:SMPLX,bib:star,bib:gdna}, face~\cite{bib:flame,hong2021headnerf}, hands~\cite{bib:MANO,li2022nimble}, and animals~\cite{bib:smal}. These models enable the transformation of mesh geometry by adjusting model parameters corresponding to pose and shape variations. Moreover, hand pose can be effectively estimated by images~\cite{li2021exploiting} or point clouds~\cite{cheng2022efficient, ren2023two}. In hand reconstruction, the parametric models like MANO~\cite{bib:MANO}, NIMBLE~\cite{li2022nimble}, have been used to recover the hand in the input image~\cite{boukhayma20193d,hasson2019learning,kong2022identityaware,cao2021reconstructing,doosti2020hopenet,hasson2020leveraging}. In~\cite{fan2021learning, Ren2023EndtoEndWS, Kim_2021_ICCV, 9710320, 9880324}, the parametric models are employed to reconstruct two hands, where hand interactions and gestures could be simulated. Recently, ~\cite{chen2023tracking} employ the HandTrackNet to track the variations of MANO parameters, which is utilized for hand-object interactions. While parametric methods are able to recover hand poses and shapes, the resulting meshes lack the capability to represent geometric textures.

\subsection{Model-Free Hand Reconstruction}
Parametric models~\cite{bib:MANO} are valuable for incorporating the prior knowledge of pose and shape, while their representation power is constrained by imposing the template shape and details. To overcome this limitation, various approaches have been explored. Instead of directly regressing the MANO parameters, I2L-MeshNet~\cite{Moon_2020_ECCV_I2L-MeshNet} predicts a 1D heatmap for each vertex coordinate, and~\cite{ge2019handshapepose, 9156522} employ GCN-based methods to recover hand meshes. On the other hand, DeepHandMesh~\cite{Moon_2020_ECCV_DeepHandMesh} make use of an encoder-decoder framework to generate highly detailed hand meshes. To achieve photo-realistic hand rendering, HARP~\cite{karunratanakul2022harp} suggest an optimization-based approach to recover both normal and albedo maps. Recently,~\cite{Luan_2023_CVPR} introduce a frequency decomposition loss to capture personalized hand from a single image, which address the problem of data scarcity through multi-view reshaping. HandAvatar~\cite{bib:handavatar} yields occupancy and illumination field to generate free-pose photo-realistic hand avatar.% Due to the limitations imposed by learning-based approaches and scarce datasets, these methods have difficulties in representing fine-grained mesh details.

\subsection{Neural Rendering-Based Reconstruction}

In past few years, rapid progress in 3D modeling and image synthesis has been obtained through neural implicit representations~\cite{occupancy_net,NeRF}. In contrast to the classical discrete representations like meshes, point clouds, and voxels, neural implicit representations leverage neural networks to model scenes, which offer continuous results with higher fidelity and flexibility. Among the various methods, Neural Radiance Fields (NeRF)~\cite{NeRF} gains great popularity and demonstrates impressive performance across various tasks. Nevertheless, the traditional NeRF-based methods face challenges in dealing with objects having temporal changes~\cite{Fu2022GeoNeus, wang2021neus}.

Neural Body~\cite{peng2021neural} combines NeRF with the parametric SMPL body model~\cite{bib:SMPL}, which is able to recover dynamic objects. Moreover,~\cite{liu2021neural} leverage NeRF to learn pose-related geometric deformations and textures in canonical space from multi-view videos. Recently, LISA ~\cite{bib:LISA} fuses volumetric rendering with hand geometric priors to capture animatable hand appearances. HandNeRF~\cite{handnerf} represents the interactive hands by deformable neural radiance fields to generate photo-realistic images.  % SHERF~\cite{SHERF} introduced a 3D perception hierarchy feature library capable of learning and rendering from a single image.

%-------------------------------------------------------------------------
\section{Method}
\label{sec:method}
Our objective is to achieve fine-grained 3D hand mesh reconstruction and synthesize photo-realistic novel views. To tackle this challenge, our method consists of three key steps. Firstly, we estimate an initial coarse hand mesh and the parameters of MANO model~\cite{bib:MANO}, which are recovered from multi-view images through incorporating multi-layer perceptrons (MLPs) into GCN. Secondly, an HAM optimization module is introduced to restore a fine-grained mesh with the albedo map and surface details by leveraging the power of inverse rendering. Finally, we suggest a mesh-based neural rendering scheme to efficiently generate photo-realistic images and refine the mesh through joint optimization. Fig.~\ref{fig:overview} illustrates the entire pipeline of our presented method.

\subsection{Coarse Initialization}
\label{sec:pose}

In hand reconstruction, it is crucial to accurately predict the 3D pose and shape due to the high flexibility of the hand. Since generating 3D poses is typically challenging in  the wild cases, some approaches like HandAvatar~\cite{bib:handavatar} employ the annotated hand poses and shapes as initialization. Instead, we aim to estimate the consistent 3D hand poses from multi-view images.

\begin{figure}
	\centering
        \includegraphics[width=0.45\textwidth]{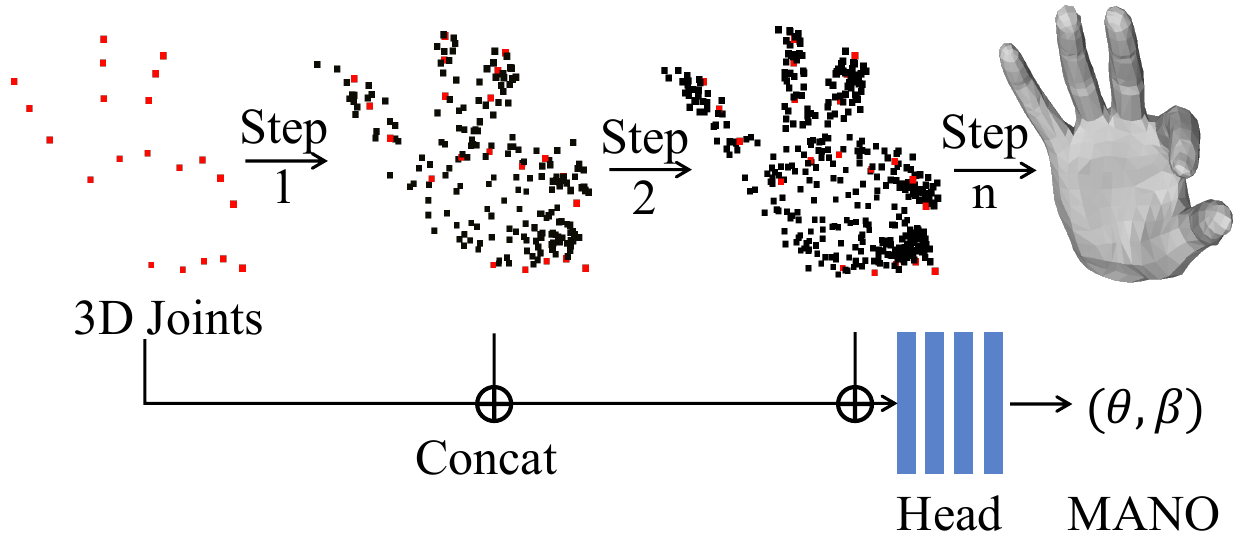}
        %\vspace{-0.1in}
	\caption{Our GCN-based network. The four-layer GCN progressively doubles the number of vertices and MANO head outputs the corresponding MANO parameters.}
	\label{fig:gcn_net}
	%\vspace{-0.2in}
\end{figure}

Given a set of images $\mathcal{I} = \{I^1, \cdots, I^n\}$ captured by the calibrated cameras and their corresponding 2D joints $\mathcal{J}_{2D} = \{J_{2D}^1, \cdots,  J_{2D}^n\}$, 3D hand joints $J_{3D} \in \mathbb{R}^{B\times3}$ in world space with $B$ per-bone parts can be estimated. To address the limitations of representing joint Euler angles directly with 3D joints information, we introduce a GCN-based network $\mathcal{G}$ to recover the MANO model as in~\cite{bib:pose2mesh}.  \begin{equation}
    \mathcal{M}(\hat{\theta}, \hat{\beta})  = \mathcal{G}(J_{3D}),
\end{equation}where $\hat{\theta} \in \mathbb{R}^{B\times3}$ and $\hat{\beta} \in \mathbb{R}^{10}$ represent pose and shape parameters of MANO model, respectively. The network $\mathcal{G}$ is designed as a four-layer GCN with a MANO head. The MANO head consists of MLPs to obtain the corresponding MANO parameters via the features of the first three GCN layers, as shown in Fig.~\ref{fig:gcn_net}. % with the input being the output of the first three stages of the GCN.

Given $n$-view images $I^n$ along with their corresponding camera parameters $\pi^n$ and the positions of the hand joints $J_{2D}^n$ in the images, $J_{3D}$ can be obtained by minimizing the reprojection error across different views. The camera parameters include the intrinsic matrix $K$ and extrinsic matrix $T$. In the $i$-th view, $J_{2D}^i$ can be calculated by $J_{2D}^i = \pi^i(J_{3D})$. Since it is difficult to circumvent the issue of imperfect estimation of $J_{2D}$, we recover $J_{3D}$ by minimizing the following objective
\begin{equation}
    \mathcal{L}_{joints} = \sum_{i=1}^{n}\frac{1}{n}||J_{2D}^i - \pi^i(\hat{J}_{3D})||^2.
\end{equation} 

Once the 3D joints $\hat{J}_{3D}$ with multi-view consistency is obtained, the GCN-based network $\mathcal{G}$ can be used to recover the MANO parameters. $\mathcal{G}$ is trained using the annotated datasets by minimizing the following energy function
\begin{equation}
    \mathcal{L}_{\mathcal{G}} = \mathcal{L}_{v} + \mathcal{L}_{n} + \mathcal{L}_{j} + \mathcal{L}_{MANO}.
\end{equation}
Specifically, $\mathcal{L}_{v}$ measures the $L_1$ distance between the predicted vertices $\hat{V}$ and annotated MANO vertices $V$ as below\begin{equation}
    \mathcal{L}_{v}  = \sum|\hat{V} - V|.
\end{equation} $\mathcal{L}_{n}$ ensures the alignment of mesh normals $\mathbf{n}$ with MANO, which is defined as \begin{equation}
    \mathcal{L}_{n} = \sum|\hat{\mathbf{n}} - \mathbf{n}|.
\end{equation} $\mathcal{L}_{j}$ constrains the discrepancy between the generated MANO joints $\hat{J_{3D}}$ and the input joints $J_{3D}$ as follows \begin{equation}
     \mathcal{L}_{j} = \sum|\hat{J_{3D}} - J_{3D}|.
\end{equation} Additionally, $\mathcal{L}_{MANO}$ is used to supervise the generation of MANO parameters, which is formulated as below \begin{equation}
    \mathcal{L}_{MANO} = \sum|(\hat{\theta}, \hat{\beta}) - (\theta, \beta))|.
\end{equation}
While the GCN-based network $\mathcal{G}$ excels at accurately recovering the pose by utilizing 3D joints as input, it lacks information about hand fatness or leanness. To overcome this limitation, we incorporate segmentation map obtained by promptable SAM~\cite{kirillov2023segment} in order to optimize the shape parameters $\hat{\beta}$. This extra optimization step further enhances the accuracy of the MANO parameters.  % and also enforce regularization on edge lengths to ensure a uniform vertex distribution

\subsection{Mesh Refinement}
\label{sec:sfs}

The MANO model generated by $\mathcal{G}$ represents a smoothing mesh without geometric textures. Inspired by the Shape from Shading (SFS) algorithm~\cite{sfs} and PatchShading~\cite{bib:nccsfs, lin2022multiview}, we accurately capture the folds and textures on the coarse hand mesh. Assuming that the diffuse reflection of human skin follows Lambertian reflectance, we propose a Hand Albedo and Mesh (HAM) optimization module to refine the coarse mesh, which takes advantage of both SFS and inverse rendering. To ensure the coherence in the mesh optimization process, we incorporate four effective regularization terms described in the following.

We utilize $\mathcal{M}_r$ as the initial model that is obtained from the MANO subdivision with 49,281 vertices and 98,432 faces. The albedo values $\rho$ are assigned to each vertex and an inverse renderer $\zeta$ is employed to generate albedo map $\mathbf{C}_{\rho}$ and normal map $\mathbf{C}_{N}$. The inverse renderer $\zeta$ is represented as follows\begin{equation}
\mathbf{C} = \zeta(c,V,F,\pi) \label{eq:render},
\end{equation} where $V$ and $F$ denote the vertices and faces of mesh $\mathcal{M}_r$, respectively. The feature at each vertex, such as normal and albedo, is represented by $c$. The rendering map $\mathbf{C}$ is obtained by the inverse renderer $\zeta$. With the calibrated images $I^i$, the HAM optimizes the vertex positions $V$ and vertex albedo $\rho$ by minimizing the texture loss $\mathcal{L}_{t}$ in the following equation \begin{equation}
\mathcal{L}_{t} = \sum\limits_{n}|B(\pi^i)  - I^i|,
\end{equation} where $B(\pi^i)$ represents the rendered image under camera parameters $\pi^i$. It is obtained by computing the illumination matrix $\mathbf{G}$, the normal map $\mathbf{C}_N$ and the albedo map $\mathbf{C}_\rho$ as below 
\begin{equation}
B(\pi^i) = \mathbf{C}_\rho\cdot SH(\mathbf{G}, \mathbf{C}_N),
\end{equation} where $SH(\cdot)$ represents sphere harmonic (SH) function with the third order. Considering the variations in lighting across different views, we optimize the lighting matrix $\mathbf{G}$ during the process. We compute the texture loss between $B(\pi^i)$ and the original image $I^i$ as the loss $\mathcal{L}_{t}$.

% In cases where both the vertex positions $V$ and vertex albedo $\rho$ are learnable, we have devised an optimization method that simultaneously optimizes these variables.

To enhance the efficacy of extracting geometric information from shadows, the regularization term $\mathcal{L}_{\rho}$ is introduced. This term is especially designed to align with the observation that human skin tends to exhibit the consistent color. $\mathcal{L}_{\rho}$ is defined as follows\begin{equation}
\mathcal{L}_{\rho} = \lambda_{1}L * \rho,
\end{equation}where $L$ denotes the Laplacian matrix. $\lambda$ is the balanced weight. To ensure the optimized vertices to be smoothing, we introduce the following term for conforming to the geometric characteristics of the hand\begin{equation} \begin{aligned}
\mathcal{L}_{r} &= \lambda_{2}L * V + \lambda_{3}\mathcal{L}_{mask} + \lambda_{4}\mathcal{L}_{e} + \lambda_{5}\mathcal{L}_{d} \\ &= \lambda_{2}L * V + \lambda_{3}\sum|\hat{M} - M_{MANO}| \\ &+ \lambda_{4}\sum\limits_{i,j}||E_{ij}||^2 + \lambda_{5}\sum\limits_{i}||\triangle{V_i}||^2. \label{eq:eq1}
\end{aligned}\end{equation}Specifically, $\mathcal{L}_{mask}$ represents the $L_1$ loss between the rendered mask and the original MANO mask. $E_{ij}$ is obtained by calculating the Euclidean distance $||\cdot||^2$ between adjacent vertices $V_i$ and $V_j$ on the mesh edges. $\mathcal{L}_{e}$ is employed to restrict the length of $E_{ij}$. Let $\triangle{V_i}$ represent the displacement distance of vertices $V_i$. $\mathcal{L}_{d}$ is utilized to ensure that the optimized hand remains close to the MANO model. Each term is assigned with a constant coefficient denoted by $\lambda$. The overall loss function $\mathcal{L}_{total}$ is defined as below \begin{equation}
\mathcal{L}_{total} = \mathcal{L}_{t} + \mathcal{L}_{\rho} + \mathcal{L}_{r}.
\end{equation}

% Specifically, $\mathcal{L}_{mask}$ represents the $L_1$ loss between the rendered mask and the original MANO mask, which is formulated as below\begin{equation}
% \mathcal{L}_{mask} = \sum|\hat{M} - M_{MANO}|.
% \end{equation}$\mathcal{L}_{e}$ is employed to restrict the length of $E_{ij}$ between $V_i$ and $V_j$, which is defined as \begin{equation}
% \mathcal{L}_{e} = \sum\limits_{i,j}||E_{ij}||^2  \label{eq:edge}.
% \end{equation} Let $\triangle{V_i}$ represent the displacement distance of vertices $V_i$. $\mathcal{L}_{d}$ is utilized to ensure that the optimized hand remains close to the MANO model as follows \begin{equation}
% \mathcal{L}_{d} = \sum\limits_{i}||\triangle{V_i}||^2  \label{eq:delta}.
% \end{equation}  Each term is assigned a weight denoted by $\lambda$. The overall loss function is defined as below \begin{equation}
% \mathcal{L}_{total} = \mathcal{L}_{t} + \mathcal{L}_{\rho} + \mathcal{L}_{r}.
% \end{equation}

The HAM module facilitates the refinement of the hand model by jointly optimizing both the mesh vertices $\hat{V}$ and albedo $\hat{\rho}$. This results in a high-quality output $\mathcal{M}_{f}$ that retains the consistent topology.
%-------------------------------------------------------------------------

\subsection{Texture Optimization and Joint Refinement} %Mesh-based Nerual Rendering
\label{sec:neuralrender}

While having achieved geometric alignment of the fine mesh, there are still some limitations in image rendering and mesh refinement. To alleviate this issue, we adopt a mesh-based neural rendering method. Leveraging the preserved topology of our refined hands, the neural rendering model can be pre-trained with diverse hand data, thereby reducing the training time for each individual hand. We propose an efficient strategy that involves the pre-training a neural rendering network using a large amount of hand data, followed by fine-tuning on individual data, and ultimately conducting joint optimization with the mesh to achieve hyper-realistic rendering and accurate geometry. % In contrast to traditional methods, neural rendering approaches~\cite{handnerf, wang2021neus} offer enhanced realism, albeit often requiring longer training times.  

% However, considering our existing finely detailed meshes, we have opted for a mesh-based neural rendering approach. This approach eliminates the need for additional point sampling and enables precise image rendering based on vertex features directly on the surface.

% Furthermore, leveraging the fixed topology structure of the subdivided MANO model, hand meshes in various poses can exhibit consistent features. To expedite the neural rendering training process, we undertake pre-training of a neural rendering network using diverse data encompassing different poses and textures. This pre-trained network significantly accelerates subsequent neural rendering training procedures.

Having acquired a refined mesh denoted as $\mathcal{M}_{f}(\hat{V}, F)$ along with its corresponding vertex albedo $\hat{\rho}$, we aim to enhance the accuracy of the rendered texture. To this end, we design a neural renderer $\mathcal{T}$ defined as below 
\begin{equation}
    \mathbf{t}(\mathbf{r}) = \mathcal{T}(\mathbf{x}, \mathbf{f}, \rho, \mathbf{n}),
\end{equation} 
where the output pixel value $\mathbf{t}$ is determined by the direction of the light ray $\mathbf{r}$. At a given pixel position, the texture field $t$ is adjusted with respect to the position $\mathbf{x}$, feature vector $\mathbf{f}$, albedo $\rho$, and normal $\mathbf{n}$, which are obtained by inverse rendering $\zeta$. For the neural renderer, we employ the MLPs with four layers, each of which consists of 256 dimensions. The length of vertex features $\mathbf{f}$ is set to 20. The training loss is defined as follows \begin{equation}
    \mathcal{L}_{tex}=\sum|\hat{\mathbf{t}}_i - I_i|.
\end{equation} 

% Compared to U-net~\cite{unet}, MLPs offer faster inference speed without compromising accuracy in this task.

\noindent\textbf{Pre-training.} During the pre-training, we conduct sampling from the subdivided MANO mesh $\mathcal{M}_r$ instead of utilizing $\mathcal{M}_f$. Through the pre-training, we obtain a well-trained neural renderer that unifies the vertices features $\mathbf{f}$ independent from position $\mathbf{x}$, albedo $\rho$ and normal $\mathbf{n}$. % To train the pre-trained model, we employ the Interhand2.6M~\cite{Moon_2020_ECCV_InterHand2.6M} that consists of images with diverse poses and textures along with their corresponding MANO models. 

\noindent\textbf{Fine-tuning.} Given the diversity of data and potential overfitting during training, the pre-trained neural rendering model often yields average results, which is unable to capture intricate texture details. To synthesize high-fidelity hand images and achieve promising rendering quality, it is necessary to fine-tune the model for each specific dataset. The advantage of the pre-trained model lies in its capacity to accelerate neural rendering, allowing for efficient completion of fine-tuning within minutes. During the fine-tuning process, we employ the vertices of mesh $\mathcal{M}_f$ obtained from Mesh Refinement along with its corresponding albedo for training. Both the vertex features $\mathbf{f}$ and the neural renderer $\mathcal{T}$ are set to be learnable.

\noindent\textbf{Joint Optimization.}
% \subsection{Joint Optimization} 
% \label{sec:jointoptim}
After achieving fine geometric structures and realistic image rendering, it becomes necessary to perform joint optimization on both mesh and texture to further enhance the overall quality. Drawing inspiration from \cite{walker2023explicit}, we adopt a geometry-based shader $\tilde{\mathbf{t}}$ with the detached output $\hat{\mathbf{t}}$ of $\mathbf{t}(\mathbf{r})$, which is illustrated as follows \begin{equation}
     \tilde{\mathbf{t}}(\mathbf{r}) = \tilde{\mathcal{T}}(\hat{\mathbf{t}}, f, \rho, \mathbf{n}).
\end{equation} 

To fine-tune the geometry and train the geometry-based shader, the loss functions $\mathcal{L}_{e}$ and $\mathcal{L}_{d}$ depicted in Eq.~\ref{eq:eq1} are employed to ensure surface smoothness. Moreover, the vertices are designated to be learnable. The shader loss is formulated as follows \begin{equation}
    \mathcal{L}_{geo}=\sum|\tilde{\mathbf{t}}_i - I_i| + \gamma_1\mathcal{L}_{e} + \gamma_2\mathcal{L}_{d}.
\end{equation}
where $\gamma_1$ and $\gamma_2$ are weights for ${L}_{e}$ and ${L}_{d}$, respectively. 

%Details: Regarding the implementation details, we utilize U-Net~\cite{unet} as our neural renderer instead of Multilayer Perceptron (MLP). This choice is motivated by U-Net's ability to incorporate contextual information, resulting in unambiguous rendered results. The feature maps fed into the U-Net are derived from Eq.~\ref{eq:render}. During the pre-training phase, we sample from the subdivided MANO mesh $\mathcal{M}_r$ instead of utilizing the optimized model obtained in Section~\ref{sec:sfs}. Due to the diversity of data and overfitting during training, the results generated by pre-trained neural rendering models tend to be an average result, which cannot show texture details. To obtain more accurate rendering images, fine-tuning is necessary for each data using the mesh $\mathcal{M}_{f}$. Pre-trained models are advantageous in accelerating neural rendering to the extent that fine-tuning can be completed within minutes.

%------------------------------------------------------------------------
\section{Experiments}
\label{sec:experiment}

% In this section, we firstly introduce the testbeds in our experiments. Then, we provide the implementation details and evaluation metrics. Finally, we present and discuss the overall experimental results of our proposed method.

%-------------------------------------------------------------------------
\subsection{Datasets}
\label{sec:datasets}
\noindent\textbf{InterHand2.6M.} InterHand2.6M~\cite{Moon_2020_ECCV_InterHand2.6M} is a large-scale dataset, comprising images of size $512\times334$ pixels and associated MANO annotations. The dataset contains multi-view temporal sequences of single hand as well as interacting hands. Our experiments primarily focus on the 5 FPS version of the InterHand2.6M dataset. In all experiments, both the GCN-based network and pre-trained neural rendering network are trained using the data from the training set. %, and we selected 10 views from the test set for mesh optimization and fine-tuning of neural rendering, while remaining views were used to validate the rendered results.

\noindent\textbf{DeepHandMesh.} The DeepHandMesh dataset~\cite{Moon_2020_ECCV_DeepHandMesh} consists of images captured from five different views with the same size as the images in InterHand2.6M dataset. Additionally, this dataset provides the corresponding 3D hand scans, which enables the validation of the mesh reconstruction quality against 3D ground truth.

\noindent\textbf{Our Dataset.} Due to the restricted resolution of the above datasets, the attainment of higher geometric precision and color fidelity requires high-resolution hand images. To address this issue, we collect a dataset using 16 calibrated cameras, which captures the synchronized images at a resolution of $1280\times1024$ pixels at 15 FPS. The cameras are distributed mainly in a semi-circle and placed at various heights to ensure a comprehensive visual coverage. 

% \begin{figure*}
%     \centering
%     \includegraphics[width=1 \linewidth]{fig/repose.pdf}
%     \caption{\textbf{Novel pose Rendering}. Our method is capable of transforming the hand to any desired pose using linear blend skinning (LBS).}
%     \label{fig:repose}
%     %\vspace{-0.1in}
% \end{figure*}

\begin{figure*}
    \centering
    \includegraphics[width=0.9 \linewidth]{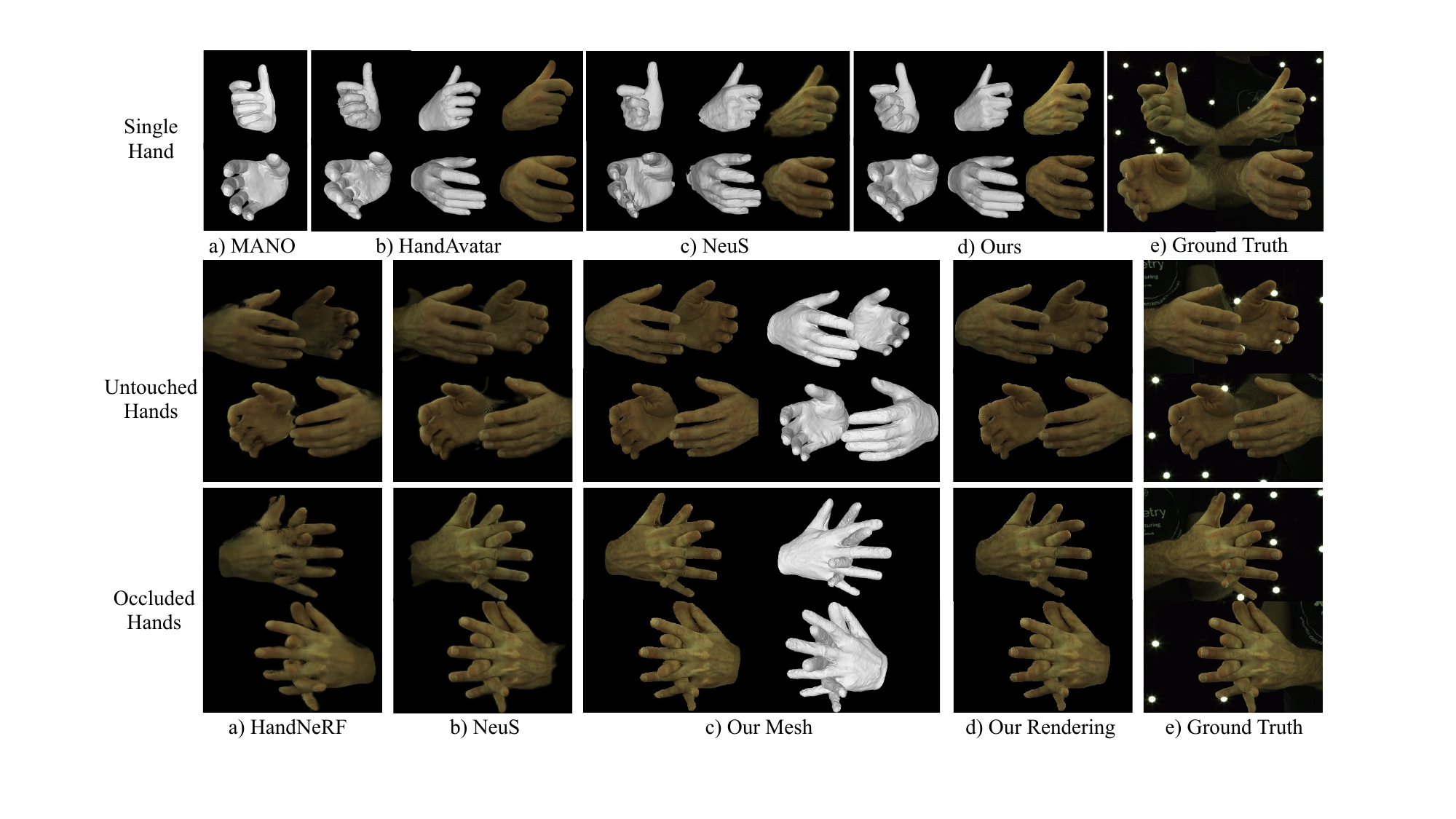}
    %\vspace{-0.25in}
    \caption{Qualitative performance comparison. We show the rendering results of single hand (first two rows) and dual hands (last four rows), which are optimized and trained from 10-view images. The hands rendered with pure white color represent the shading in order to highlight the level of mesh detail.}
    \label{fig:vis}
    %\vspace{-0.1in}
\end{figure*}

\begin{table*}[t]
\small
\centering
\begin{tabular}{c || c c c || c c c || c c c}
\hline
\multirow{2}{*}{Method} & \multicolumn{3}{c||}{\textit{test/Capture0-Single}} & \multicolumn{3}{c||}{\textit{test/Capture1-Single}}& \multicolumn{3}{c}{\textit{test/Capture0-Interacting}} \\ 
 & LPIPS $\downarrow$ & PSNR $\uparrow$ & SSIM $\uparrow$ & LPIPS $\downarrow$ & PSNR $\uparrow$ & SSIM $\uparrow$ & LPIPS $\downarrow$ & PSNR $\uparrow$ & SSIM $\uparrow$  \\
\hline 

Ani-NeRF & 0.0621 & 31.78 & 0.968 & - &  - & -  & 0.0798  & 29.35  &  0.949 \\
HandAvatar &  0.0504 & 33.01  &  0.933 & 0.0425  &  32.12 & 0.938  & - & - & -  \\
HandNerf & 0.0375 & 32.70 & 0.974 & - &  - & -  & 0.0367  & 30.62  &  0.958 \\
NeuS & 0.0197 & 35.34 & 0.986 & 0.0211 & 34.98 & 0.984  &  0.0743 &  31.46 & 0.927 \\
\hline
Ours & \bf0.0069 & \bf37.26 & \bf0.992 & \bf0.0098 & \bf37.91 & \bf0.991  & \bf0.0126  & \bf37.08  & \bf0.986 \\
\hline
\end{tabular}
%\vspace{-0.2cm}
\caption{Rendering quality comparisons among our method and prior arts on the InterHand2.6M dataset.}
\label{tab:render}
%\vspace{-0.45cm}
\end{table*}

%-------------------------------------------------------------------------
\subsection{Implementation Details and Metrics}
\label{sec:detail}
% \noindent\textbf{Pre-training} We conducte pre-training of the GCN and neural rendering networks using the InterHand2.6M dataset~\cite{Moon_2020_ECCV_InterHand2.6M}. The GCN network comprise four layers, with the number of vertices doubling in each layer. 

\noindent\textbf{Optimization.} To achieve the fine-grained mesh, we adopt a subdivision technique inspired by \cite{bib:handavatar}, which expands the original 778 vertices in the MANO model to a total of 49,281 vertices. During the optimization process, we utilize the Adam optimizer~\cite{Adam} with the balanced weights of $\lambda_1=20$, $\lambda_2=40$, $\lambda_3=20$, $\lambda_4=100$, and $\lambda_5=2$ to jointly optimize the vertices, vertex albedo, and lighting coefficients over 100 iterations. This optimization process takes approximately 20 seconds. Additionally, the neural renderer is pre-trained on InterHand2.6M dataset for 20 epochs. Subsequently, for fine-tuning and joint optimization, each process requires 100 epochs of training with $\gamma_1=100$ and $\gamma_2=2$, respectively. Notably, the entire optimization pipeline is computationally efficient, which takes approximately 90 seconds on a single NVIDIA 3090Ti GPU.

\noindent\textbf{Evaluation Metrics.} We evaluate the accuracy of the reconstructed 3D surface by computing the average point-to-surface Euclidean distance (P2S) between the vertices of the recovered surface and their corresponding ground truth, which are measured in millimetre. Due to the disparate size ranging between the generated hand mesh and the 3D scans in the DeepHandMesh data, the Chamfer distance metric is considered unsuitable. In line with prior research in neural rendering, peak signal-to-noise ratio (PSNR), structural similarity index (SSIM), and learned perceptual image patch similarity (LPIPS) are adopted as evaluation metrics to gauge the fidelity of the synthesized image results.

%-------------------------------------------------------------------------
\subsection{Results on Hand Reconstruction}  % Shape and Color Reconstruction from Images
\label{sec:exp}
% Table~\ref{tab:render} and Table~\ref{tab:mesh} summarize the performance metrics of ours and baselines in terms of image rendering and geometry. Fig~\ref{fig:vis} and~\ref{fig:mesh} showcase the visual results. The results of our dataset are presented in Fig~\ref{fig:ourdata}. Additional results can be found in the supplementary material.

%The results of InterHand2.6M dataset are reported in Fig.~\ref{fig:vis} and Table~\ref{tab:render}. Fig.~\ref{fig:mesh} and Table~\ref{tab:mesh} showcase the quality of hand reconstruction in DeepHandMesh dataset. The results of our dataset are presented in Fig.~\ref{fig:ourdata}. Additional results can be found in the supplementary material.

\begin{table}[t]
\centering
	\begin{tabular}{c| c c c | c}
            \hline
            \multirow{1}{*}{P2S $\downarrow$} & \makecell{\textit{Two} \\ \textit{occlusions}} & \makecell{\textit{Thumb} \\ \textit{tuck}} & \makecell{\textit{Shake-} \\ \textit{speare}} & \makecell{\textit{Total} \\ \textit{Average}} \\
            \hline 
		DHM & 3.467 & 2.624 & 1.773 & 2.937 \\
            w/o.HAM &  1.637 & 2.317 & 1.668 & 1.873 \\
	    Ours &  \bf1.532 & \bf1.281 & \bf1.271 & \bf1.456 \\
            \hline
	\end{tabular}
 %\vspace{-0.05in}
\caption{Mesh reconstruction quality comparison among ours and DHM~\cite{Moon_2020_ECCV_DeepHandMesh} on DeepHandMesh dataset with 5 views.}
\label{tab:mesh}
 %\vspace{-0.25in}
\end{table}

% \begin{table*}[t]
% %\renewcommand{\arraystretch}{1.}
% % \setlength\tabcolsep{2pt}
% \centering
% 	\begin{tabular}{c | c c c | c c c}
%             \hline
%             \multirow{2}{*}{Method} & \multicolumn{3}{c|}{\textit{test/Capture0}} & \multicolumn{3}{c}{\textit{test/Capture1}} \\
%              & LPIPS $\downarrow$ & PSNR $\uparrow$ & SSIM $\uparrow$ & LPIPS $\downarrow$ & PSNR $\uparrow$ & SSIM $\uparrow$ \\
%             \hline 
%             HandAvatar~\cite{bib:handavatar} & 0.0473 & \bf31.04 & 0.953 & 0.0515 & 29.06 & 0.947 \\
% 		  Ours & \bf0.0131 & 30.54 & \bf0.972 & \bf0.0098 & \bf30.94 & \bf0.965 \\
%             \hline
% 	\end{tabular}
% \caption{Novel poses rendering quality among ours and HandAvatar~\cite{bib:handavatar} on InterHand2.6M~\cite{Moon_2020_ECCV_InterHand2.6M} dataset.}
% \label{tab:repose}
%  %\vspace{-0.1in}
% \end{table*}

\noindent\textbf{Results of InterHand2.6M.} To evaluate the quality of novel view synthesis, we conduct experiments on the InterHand2.6M dataset using 10 views and evaluate on the rest views. Table~\ref{tab:render} summarizes the evaluation results on rendering quality. Fig.~\ref{fig:vis} illustrates the visual comparisons against HandNeRF~\cite{handnerf}, HandAvatar~\cite{bib:handavatar} and NeuS~\cite{wang2021neus}. The metrics of Ani-NeRF are extracted from the data presented in~\cite{handnerf}. It is important to note that HandAvatar lacks support for interactive hands, while HandNeRF is not able to directly predict geometry. Both HandNeRF and HandAvatar rely on learning from video sequence for voxel rendering with large pose variations, which may result in smoothing texture. By taking advantage of the design of our topology-consistent hand mesh and the mesh-based neural rendering network, our presented method achieves PSNR of 37dB with just about one minute of fine-tuning. Comparing to NeuS, some semi-transparent mist-like artifacts are observed around the rendered hand, as shown in Fig.~\ref{fig:vis}.
% Notably, our approach is applicable to any number of hands, although variations in the hand count may result in changes in the vertex count of the mesh. However, the topological consistency ensures that the features of each vertex remain fixed.

\noindent\textbf{Results of DeepHandMesh.} In the DeepHandMesh dataset, five views are utilized to estimate MANO parameters and optimize the mesh. 3D scans of DeepHandMesh is employed to assess the geometric quality. Table~\ref{tab:mesh} reports the mesh geometry quality, and Fig.~\ref{fig:mesh} shows the generated mesh. Unlike DeepHandMesh relying on weakly supervised learning from depth maps, our method leverages the diffuse reflection assumption and incorporates the HAM module to achieve fine-grained hand reconstruction with sparse views. It is challenging for learning-based methods to capture the personalized details on DeepHandMesh, which leads to smoothing results due to their generalization capability. In contrast, the multi-view reconstruction method, NeuS~\cite{wang2021neus}, fails due to the insufficient number of views. % As the availability of scanning data for reconstructing highly detailed hand models is limited, methods that excel in this area are scarce.

\label{sec:abl}
\begin{table}
    \centering
	\begin{tabular}{c c c | c c}
            \toprule
            \textit{Init} & \textit{HAM} & \textit{Joint refine} & {P2S $\downarrow$} & {PSNR $\uparrow$} \\
            \hline 
            \checkmark &  &  & 1.873 & - \\
		  & \checkmark & \checkmark & 5.506 & 31.96 \\
    	\checkmark & \checkmark &  & 1.457 & 35.68 \\
             \checkmark  & \checkmark & \checkmark & \bf1.456 & \bf37.91 \\
            \bottomrule
	\end{tabular}
  %\vspace{-0.05in}
\caption{The impact of coarse initialization, HAM and joint refinement on rendering quality and mesh geometry. }
    %\vspace{-0.25in}
 \label{tab:ablation}
\end{table}

\begin{figure}
	\centering
        \includegraphics[width=0.45\textwidth]{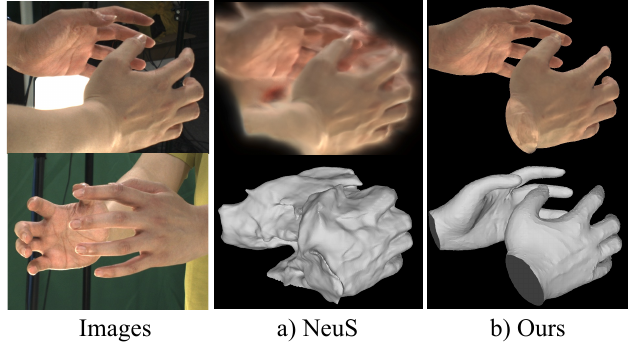} % ourdataset.pdf
        %\vspace{-0.1in}
	\caption{Results of our dataset.  To evaluate both geometric accuracy and rendering quality, we compare our framework with NeuS~\cite{wang2021neus} on our dataset.}
	\label{fig:ourdata}
	%\vspace{-0.15in}
\end{figure}

\noindent\textbf{Results of Our Dataset.} Fig.~\ref{fig:ourdata} shows the results of our dataset. Our dataset exhibits a more authentic color quality. The experimental results indicate that our proposed approach is effective for various datasets with different capture devices. Due to the absence of hand priors, the mesh generated by NeuS~\cite{wang2021neus} may not conform to the characteristics typically associated with hands. % Additional results can be found in the supplementary material.

% \noindent\textbf{Novel Pose Synthesis} After obtaining a detailed mesh and rendering network for a specific pose, our method utilizes MANO parameters and linear blending skinning (LBS) algorithms to transform the mesh to novel poses in other sequences. This process is summarized in Table \ref{tab:repose}. Experimental results demonstrate that our approach can accurately render hands in novel poses by transforming the mesh. In comparison, HandAvatar~\cite{bib:handavatar} employs video sequences for training, resulting in more realistic hand lighting variations across poses, while the color texture tends to appear smoother. While PSNR reflects the degree of image distortion, LPIPS is more meaningful~\cite{handnerf}. Conversely, NeuS~\cite{wang2021neus} lacks a fixed topology structure and is hard to perform pose transformations.

%-------------------------------------------------------------------------
\subsection{Ablation Study}

% We conducted ablation experiments to validate the effectiveness of the three components of HEIR. The results are presented in Table \ref{tab:ablation} and Figures \ref{fig:gcn} and \ref{fig:ablation}.

\begin{figure}
	\centering
    \resizebox{\linewidth}{!}{
        \includegraphics[width=0.45\textwidth]{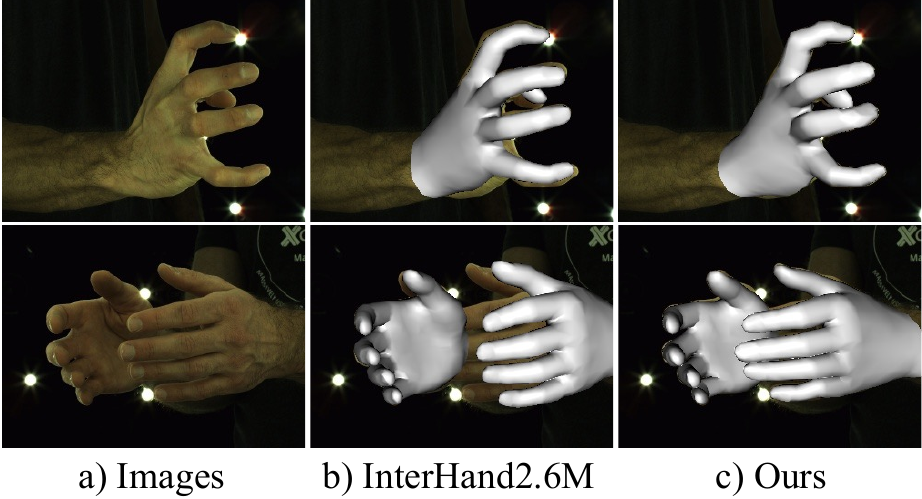}}
        %\vspace{-0.25in}
	\caption{Comparison on MANO meshes. We compare our proposed GCN-based network with the annotations from the InterHand2.6M dataset.}
	\label{fig:gcn}
	%\vspace{-0.05in}
\end{figure}

\noindent\textbf{Coarse Initialization.} We directly compare our results with the annotations in the InterHand2.6M dataset. As illustrated in Fig.~\ref{fig:gcn}, the MANO model generated by our GCN-based network exhibits a closer alignment with the hands in the images. Moreover, Table~\ref{tab:ablation} demonstrates that directly using the MANO parameters provided by the dataset may introduce errors, which will subsequently affect the rendering quality and reconstruction results. % It should be noted that the manual annotations in the InterHand2.6M dataset have limitations in terms of representing MANO parameters. While they ensure correct hand poses and approximate shapes, they do not guarantee consistency with the original hand contours.

\begin{figure}
	\centering
        \includegraphics[width=0.45\textwidth]{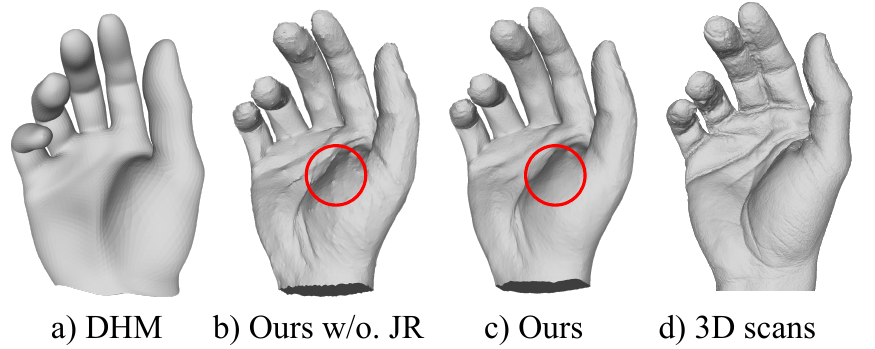}
        %\vspace{-0.1in}
	\caption{Comparison on mesh quality. The generated meshes are compared in terms of geometric quality using 5 different views on the DeepHandMesh~\cite{Moon_2020_ECCV_DeepHandMesh} dataset. JR represents the Joint Refinement.}
	\label{fig:mesh}
	%\vspace{-0.1in}
\end{figure}

\begin{figure}
	\centering
        \includegraphics[width=0.42\textwidth]{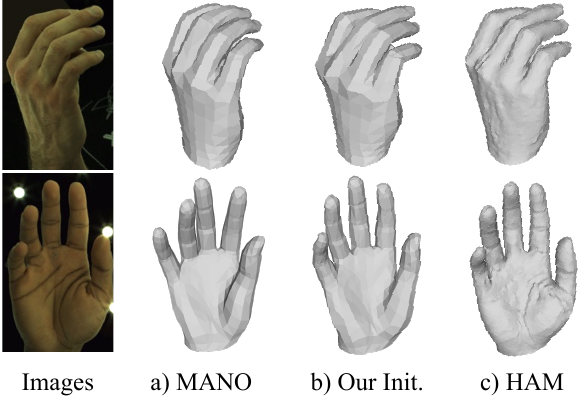}
        %\vspace{-0.1in}
        \caption{The visual results of different methods on mesh geometry quality.} % during geometric reconstruction.
	\label{fig:ablation}
	%\vspace{-0.15in}
\end{figure}

\noindent\textbf{Effects of HAM.} The results reported in Fig.~\ref{fig:ablation} provide clear evidence of the substantial impact of mesh subdivision and the HAM module in achieving fine-detailed hand reconstructions. The mesh subdivision process effectively increases the number of vertices in MANO model, while the HAM module plays a crucial role in capturing the surface wrinkles and intricate details of the reconstructed hand. As indicated in Table~\ref{tab:ablation}, the inclusion of HAM module greatly contributes to generating the surface reconstructions that closely resemble the real 3D scans.

\noindent\textbf{Texture Optimization and Joint Refinement.} Due to variations in viewpoints, the mesh generated by HAM may exhibit artifacts resulting from shading inconsistencies and ambiguous reflectance. Comparison results in Table~\ref{tab:ablation} clearly demonstrate the significant improvement in image quality achieved through joint refinement. The visualizations presented in Fig.~\ref{fig:vis} further illustrate that neural rendering excels in reconstructing accurate lighting conditions compared to the results of HAM module. Although the geometric optimization achieves the subtle metrics changes in terms of rendering results, it effectively eliminates the non-smoothing singularities, as shown in Fig.~\ref{fig:mesh}.
%Regarding geometric optimization, the limited optimization capabilities of the renderer result in subtle metrics changes. In Fig.~\ref{fig:mesh}, it can be observed that non-smoothing singularities have been effectively eliminated. 
% The visualizations depicted in Fig. \ref{fig:vis} further illustrate that neural rendering excels in reconstructing accurate lighting conditions compared to the results obtained using HAM. This superiority arises from the fact that HAM relies on assumptions regarding lighting coefficients, making it challenging to precisely represent the lighting conditions depicted in the images.

%-------------------------------------------------------------------------
\section{Conclusion}

In this paper, we introduced a novel fine-grained multi-view hand mesh reconstruction method by leveraging effective inverse rendering to restore hand poses and intricate details. Our approach predicted a parametric hand mesh model by a GCN-based network while refining both the hand mesh and textures through the Hand Albedo and Mesh (HAM) optimization module. To generate photo-realistic image, we suggested an effective mesh-based neural rendering scheme by fusing the pre-trained rendering network with vertex features. Through extensive experiments on diverse datasets, the promising results demonstrated the efficacy of our proposed approach.

\section*{Acknowledgments}
This work is supported by National Natural Science Foundation of China under Grants (62376244, 61831015). It is also supported by the Information Technology Center and State Key Lab of CAD\&CG, Zhejiang University.

\bibliography{aaai24}

\begin{thebibliography}{57}
\providecommand{\natexlab}[1]{#1}

\bibitem[{Bhatnagar et~al.(2020)Bhatnagar, Sminchisescu, Theobalt, and Pons-Moll}]{bhatnagar2020loopreg}
Bhatnagar, B.~L.; Sminchisescu, C.; Theobalt, C.; and Pons-Moll, G. 2020.
\newblock LoopReg: Self-supervised Learning of Implicit Surface Correspondences, Pose and Shape for 3D Human Mesh Registration.
\newblock In \emph{NeurIPS}, 12909--12922.

\bibitem[{Boukhayma, de~Bem, and Torr(2019)}]{boukhayma20193d}
Boukhayma, A.; de~Bem, R.; and Torr, P.~H. 2019.
\newblock 3D Hand Shape and Pose From Images in the Wild.
\newblock In \emph{CVPR}, 10835--10844.

\bibitem[{Cao et~al.(2021)Cao, Radosavovic, Kanazawa, and Malik}]{cao2021reconstructing}
Cao, Z.; Radosavovic, I.; Kanazawa, A.; and Malik, J. 2021.
\newblock Reconstructing Hand-Object Interactions in the Wild.
\newblock In \emph{ICCV}, 12397--12406.

\bibitem[{Chen et~al.(2023)Chen, Yan, Zhang, Xu, Li, Weng, Yi, Song, and Wang}]{chen2023tracking}
Chen, J.; Yan, M.; Zhang, J.; Xu, Y.; Li, X.; Weng, Y.; Yi, L.; Song, S.; and Wang, H. 2023.
\newblock Tracking and Reconstructing Hand Object Interactions from Point Cloud Sequences in the Wild.
\newblock In \emph{AAAI}, 304--312.

\bibitem[{Chen et~al.(2022)Chen, Jiang, Song, Yang, Black, Geiger, and Hilliges}]{bib:gdna}
Chen, X.; Jiang, T.; Song, J.; Yang, J.; Black, M.~J.; Geiger, A.; and Hilliges, O. 2022.
\newblock gDNA: Towards Generative Detailed Neural Avatars.
\newblock In \emph{CVPR}, 20427--20437.

\bibitem[{Chen, Wang, and Shum(2023)}]{bib:handavatar}
Chen, X.; Wang, B.; and Shum, H.-Y. 2023.
\newblock Hand Avatar: Free-Pose Hand Animation and Rendering From Monocular Video.
\newblock In \emph{CVPR}, 8683--8693.

\bibitem[{Chen et~al.(2021)Chen, Zheng, Black, Hilliges, and Geiger}]{chen2021snarf}
Chen, X.; Zheng, Y.; Black, M.~J.; Hilliges, O.; and Geiger, A. 2021.
\newblock SNARF: Differentiable Forward Skinning for Animating Non-Rigid Neural Implicit Shapes.
\newblock In \emph{ICCV}, 11594--11604.

\bibitem[{Cheng et~al.(2022)Cheng, Wan, Zuo, Ma, Gu, Tan, Wang, Deng, and Zhang}]{cheng2022efficient}
Cheng, J.; Wan, Y.; Zuo, D.; Ma, C.; Gu, J.; Tan, P.; Wang, H.; Deng, X.; and Zhang, Y. 2022.
\newblock Efficient Virtual View Selection for 3D Hand Pose Estimation.
\newblock In \emph{AAAI}, 419--426.

\bibitem[{Choi, Moon, and Lee(2020)}]{bib:pose2mesh}
Choi, H.; Moon, G.; and Lee, K.~M. 2020.
\newblock Pose2Mesh: Graph Convolutional Network for 3D Human Pose and Mesh Recovery from a 2D Human Pose.
\newblock In \emph{ECCV}, 769–787.

\bibitem[{Corona et~al.(2022)Corona, Hodan, Vo, Moreno-Noguer, Sweeney, Newcombe, and Ma}]{bib:LISA}
Corona, E.; Hodan, T.; Vo, M.; Moreno-Noguer, F.; Sweeney, C.; Newcombe, R.; and Ma, L. 2022.
\newblock LISA: Learning Implicit Shape and Appearance of Hands.
\newblock In \emph{CVPR}, 20501--20511.

\bibitem[{Doosti et~al.(2020)Doosti, Naha, Mirbagheri, and Crandall}]{doosti2020hopenet}
Doosti, B.; Naha, S.; Mirbagheri, M.; and Crandall, D.~J. 2020.
\newblock HOPE-Net: {A} Graph-Based Model for Hand-Object Pose Estimation.
\newblock In \emph{CVPR}, 6607--6616.

\bibitem[{Fan et~al.(2021)Fan, Spurr, Kocabas, Tang, Black, and Hilliges}]{fan2021learning}
Fan, Z.; Spurr, A.; Kocabas, M.; Tang, S.; Black, M.~J.; and Hilliges, O. 2021.
\newblock Learning to Disambiguate Strongly Interacting Hands via Probabilistic Per-pixel Part Segmentation.
\newblock In \emph{3DV}, 1--10.

\bibitem[{Fu et~al.(2022)Fu, Xu, Ong, and Tao}]{Fu2022GeoNeus}
Fu, Q.; Xu, Q.; Ong, Y.~S.; and Tao, W. 2022.
\newblock Geo-neus: Geometry-consistent Neural Implicit Surfaces Learning for Multi-view Reconstruction.
\newblock \emph{NeurIPS}, 3403--3416.

\bibitem[{Ge et~al.(2019)Ge, Ren, Li, Xue, Wang, Cai, and Yuan}]{ge2019handshapepose}
Ge, L.; Ren, Z.; Li, Y.; Xue, Z.; Wang, Y.; Cai, J.; and Yuan, J. 2019.
\newblock 3D Hand Shape and Pose Estimation from a Single RGB Image.
\newblock In \emph{CVPR}, 10833--10842.

\bibitem[{Grassal et~al.(2022)Grassal, Prinzler, Leistner, Rother, Nie{\ss}ner, and Thies}]{grassal2021neural}
Grassal, P.-W.; Prinzler, M.; Leistner, T.; Rother, C.; Nie{\ss}ner, M.; and Thies, J. 2022.
\newblock Neural Head Avatars from Monocular RGB Videos.
\newblock In \emph{CVPR}, 18653--18664.

\bibitem[{Guo et~al.(2023)Guo, Zhou, Wang, Li, and Li}]{handnerf}
Guo, Z.; Zhou, W.; Wang, M.; Li, L.; and Li, H. 2023.
\newblock HandNeRF: Neural Radiance Fields for Animatable Interacting Hands.
\newblock In \emph{CVPR}, 21078--21087.

\bibitem[{Hasson et~al.(2020)Hasson, Tekin, Bogo, Laptev, Pollefeys, and Schmid}]{hasson2020leveraging}
Hasson, Y.; Tekin, B.; Bogo, F.; Laptev, I.; Pollefeys, M.; and Schmid, C. 2020.
\newblock Leveraging Photometric Consistency Over Time for Sparsely Supervised Hand-object Reconstruction.
\newblock In \emph{CVPR}, 571--580.

\bibitem[{Hasson et~al.(2019)Hasson, Varol, Tzionas, Kalevatykh, Black, Laptev, and Schmid}]{hasson2019learning}
Hasson, Y.; Varol, G.; Tzionas, D.; Kalevatykh, I.; Black, M.~J.; Laptev, I.; and Schmid, C. 2019.
\newblock Learning Joint Reconstruction of Hands and Manipulated Objects.
\newblock In \emph{CVPR}, 11807--11816.

\bibitem[{Hong et~al.(2022)Hong, Peng, Xiao, Liu, and Zhang}]{hong2021headnerf}
Hong, Y.; Peng, B.; Xiao, H.; Liu, L.; and Zhang, J. 2022.
\newblock HeadNeRF: A Real-time NeRF-based Parametric Head Model.
\newblock In \emph{CVPR}, 20374--20384.

\bibitem[{Horn(1970)}]{sfs}
Horn, B. K.~P. 1970.
\newblock \emph{Shape from shading; a method for obtaining the shape of a smooth opaque object from one view}.
\newblock Ph.D. thesis, Massachusetts Institute of Technology, {USA}.

\bibitem[{Karunratanakul et~al.(2023)Karunratanakul, Prokudin, Hilliges, and Tang}]{karunratanakul2022harp}
Karunratanakul, K.; Prokudin, S.; Hilliges, O.; and Tang, S. 2023.
\newblock HARP: Personalized Hand Reconstruction from a Monocular RGB Video.
\newblock In \emph{CVPR}, 12802--12813.

\bibitem[{Kim, Kim, and Baek(2021)}]{Kim_2021_ICCV}
Kim, D.~U.; Kim, K.~I.; and Baek, S. 2021.
\newblock End-to-end Detection and Pose Estimation of Two Interacting Hands.
\newblock In \emph{ICCV}, 11189--11198.

\bibitem[{Kingma and Ba(2014)}]{Adam}
Kingma, D.~P.; and Ba, J. 2014.
\newblock Adam: A method for stochastic optimization.
\newblock \emph{arXiv preprint arXiv:1412.6980}.

\bibitem[{Kipf and Welling(2017)}]{gcn}
Kipf, T.~N.; and Welling, M. 2017.
\newblock Semi-Supervised Classification with Graph Convolutional Networks.
\newblock In \emph{ICLR}.

\bibitem[{Kirillov et~al.(2023)Kirillov, Mintun, Ravi, Mao, Rolland, Gustafson, Xiao, Whitehead, Berg, Lo, Dollar, and Girshick}]{kirillov2023segment}
Kirillov, A.; Mintun, E.; Ravi, N.; Mao, H.; Rolland, C.; Gustafson, L.; Xiao, T.; Whitehead, S.; Berg, A.~C.; Lo, W.-Y.; Dollar, P.; and Girshick, R. 2023.
\newblock Segment Anything.
\newblock In \emph{ICCV}, 4015--4026.

\bibitem[{Kong et~al.(2022)Kong, Zhang, Chen, Ma, Yan, Sun, Liu, Han, and Xie}]{kong2022identityaware}
Kong, D.; Zhang, L.; Chen, L.; Ma, H.; Yan, X.; Sun, S.; Liu, X.; Han, K.; and Xie, X. 2022.
\newblock Identity-aware Hand Mesh Estimation and Personalization from RGB Images.
\newblock In \emph{ECCV}, 536--553.

\bibitem[{Kulon et~al.(2020)Kulon, Guler, Kokkinos, Bronstein, and Zafeiriou}]{9156522}
Kulon, D.; Guler, R.~A.; Kokkinos, I.; Bronstein, M.~M.; and Zafeiriou, S. 2020.
\newblock Weakly-supervised Mesh-convolutional Hand Reconstruction in the Wild.
\newblock In \emph{CVPR}, 4990--5000.

\bibitem[{Lei et~al.(2023)Lei, Ren, Feng, Cui, and Xie}]{lei2023hierarchical}
Lei, B.; Ren, J.; Feng, M.; Cui, M.; and Xie, X. 2023.
\newblock A Hierarchical Representation Network for Accurate and Detailed Face Reconstruction from In-The-Wild Images.
\newblock In \emph{CVPR}, 394--403.

\bibitem[{Li et~al.(2022{\natexlab{a}})Li, An, Zhang, Wu, Chen, Yu, and Liu}]{9880324}
Li, M.; An, L.; Zhang, H.; Wu, L.; Chen, F.; Yu, T.; and Liu, Y. 2022{\natexlab{a}}.
\newblock Interacting Attention Graph for Single Image Two-hand Reconstruction.
\newblock In \emph{CVPR}, 2761--2770.

\bibitem[{Li, Gao, and Sang(2021)}]{li2021exploiting}
Li, M.; Gao, Y.; and Sang, N. 2021.
\newblock Exploiting Learnable Joint Groups for Hand Pose Estimation.
\newblock In \emph{AAAI}, 1921--1929.

\bibitem[{Li et~al.(2017)Li, Bolkart, Black, Li, and Romero}]{bib:flame}
Li, T.; Bolkart, T.; Black, M.~J.; Li, H.; and Romero, J. 2017.
\newblock Learning a Model of Facial Shape and Expression from 4D Scans.
\newblock \emph{ACM TOG}, 194--1.

\bibitem[{Li et~al.(2022{\natexlab{b}})Li, Zhang, Qiu, Jiang, Li, Ma, Zhang, Xu, and Yu}]{li2022nimble}
Li, Y.; Zhang, L.; Qiu, Z.; Jiang, Y.; Li, N.; Ma, Y.; Zhang, Y.; Xu, L.; and Yu, J. 2022{\natexlab{b}}.
\newblock NIMBLE: a Non-rigid Hand Model with Bones and Muscles.
\newblock \emph{ACM TOG}, 1--16.

\bibitem[{Lin et~al.(2024)Lin, Peng, Gan, and Zhu}]{bib:nccsfs}
Lin, L.; Peng, S.; Gan, Q.; and Zhu, J. 2024.
\newblock FastHuman: Reconstructing High-Quality Clothed Human in Minutes.
\newblock In \emph{3DV}.

\bibitem[{Lin, Zhu, and Zhang(2022)}]{lin2022multiview}
Lin, L.; Zhu, J.; and Zhang, Y. 2022.
\newblock Multiview textured mesh recovery by differentiable rendering.
\newblock \emph{TCSVT}, 1684--1696.

\bibitem[{Liu et~al.(2021)Liu, Habermann, Rudnev, Sarkar, Gu, and Theobalt}]{liu2021neural}
Liu, L.; Habermann, M.; Rudnev, V.; Sarkar, K.; Gu, J.; and Theobalt, C. 2021.
\newblock Neural Actor: Neural Free-view Synthesis of Human Actors with Pose Control.
\newblock \emph{ACM TOG}, 1--16.

\bibitem[{Loper et~al.(2015)Loper, Mahmood, Romero, Pons-Moll, and Black}]{bib:SMPL}
Loper, M.; Mahmood, N.; Romero, J.; Pons-Moll, G.; and Black, M.~J. 2015.
\newblock SMPL: A skinned multi-person linear model.
\newblock \emph{ACM TOG}, 1--16.

\bibitem[{Luan et~al.(2023)Luan, Zhai, Meng, Li, Chen, Xu, and Yuan}]{Luan_2023_CVPR}
Luan, T.; Zhai, Y.; Meng, J.; Li, Z.; Chen, Z.; Xu, Y.; and Yuan, J. 2023.
\newblock High Fidelity 3D Hand Shape Reconstruction via Scalable Graph Frequency Decomposition.
\newblock In \emph{CVPR}, 16795--16804.

\bibitem[{Mescheder et~al.(2019)Mescheder, Oechsle, Niemeyer, Nowozin, and Geiger}]{occupancy_net}
Mescheder, L.; Oechsle, M.; Niemeyer, M.; Nowozin, S.; and Geiger, A. 2019.
\newblock Occupancy Networks: Learning 3D Reconstruction in Function Space.
\newblock In \emph{CVPR}, 4460--4470.

\bibitem[{Mildenhall et~al.(2021)Mildenhall, Srinivasan, Tancik, Barron, Ramamoorthi, and Ng}]{NeRF}
Mildenhall, B.; Srinivasan, P.~P.; Tancik, M.; Barron, J.~T.; Ramamoorthi, R.; and Ng, R. 2021.
\newblock NeRF: Representing Scenes as Neural Radiance Fields for View Synthesis.
\newblock \emph{CACM}, 99--106.

\bibitem[{Moon and Lee(2020)}]{Moon_2020_ECCV_I2L-MeshNet}
Moon, G.; and Lee, K.~M. 2020.
\newblock I2l-meshnet: Image-to-lixel Prediction Network for Accurate 3D Human Pose and Pesh Estimation from a Single RGB Image.
\newblock In \emph{ECCV}, 752--768.

\bibitem[{Moon, Shiratori, and Lee(2020)}]{Moon_2020_ECCV_DeepHandMesh}
Moon, G.; Shiratori, T.; and Lee, K.~M. 2020.
\newblock Deephandmesh: A Weakly-supervised Deep Encoder-decoder Framework for High-fidelity Hand Mesh Modeling.
\newblock In \emph{ECCV}, 440--455.

\bibitem[{Moon et~al.(2020)Moon, Yu, Wen, Shiratori, and Lee}]{Moon_2020_ECCV_InterHand2.6M}
Moon, G.; Yu, S.-I.; Wen, H.; Shiratori, T.; and Lee, K.~M. 2020.
\newblock Interhand2.6M: A Dataset and Baseline for 3D Interacting Hand Pose Estimation from a Single RGB Image.
\newblock In \emph{ECCV}, 548--564.

\bibitem[{Noguchi et~al.(2021)Noguchi, Sun, Lin, and Harada}]{2021narf}
Noguchi, A.; Sun, X.; Lin, S.; and Harada, T. 2021.
\newblock Neural Articulated Radiance Field.
\newblock In \emph{ICCV}, 5762--5772.

\bibitem[{Osman, Bolkart, and Black(2020)}]{bib:star}
Osman, A.~A.; Bolkart, T.; and Black, M.~J. 2020.
\newblock {STAR:} Sparse Trained Articulated Human Body Regressor.
\newblock In \emph{ECCV}, 598--613.

\bibitem[{Pavlakos et~al.(2019)Pavlakos, Choutas, Ghorbani, Bolkart, Osman, Tzionas, and Black}]{bib:SMPLX}
Pavlakos, G.; Choutas, V.; Ghorbani, N.; Bolkart, T.; Osman, A.~A.; Tzionas, D.; and Black, M.~J. 2019.
\newblock Expressive Body Capture: 3D Hands, Face, and Body from a Single Image.
\newblock In \emph{CVPR}, 10975--10985.

\bibitem[{Peng et~al.(2021{\natexlab{a}})Peng, Dong, Wang, Zhang, Shuai, Zhou, and Bao}]{peng2021animatable}
Peng, S.; Dong, J.; Wang, Q.; Zhang, S.; Shuai, Q.; Zhou, X.; and Bao, H. 2021{\natexlab{a}}.
\newblock Animatable Neural Radiance Fields for Modeling Dynamic Human Bodies.
\newblock In \emph{ICCV}, 14314--14323.

\bibitem[{Peng et~al.(2021{\natexlab{b}})Peng, Zhang, Xu, Wang, Shuai, Bao, and Zhou}]{peng2021neural}
Peng, S.; Zhang, Y.; Xu, Y.; Wang, Q.; Shuai, Q.; Bao, H.; and Zhou, X. 2021{\natexlab{b}}.
\newblock Neural Body: Implicit Neural Representations with Structured Latent Codes for Novel View Synthesis of Dynamic Humans.
\newblock In \emph{CVPR}, 9054--9063.

\bibitem[{Ren, Zhu, and Zhang(2023)}]{Ren2023EndtoEndWS}
Ren, J.; Zhu, J.; and Zhang, J. 2023.
\newblock End-to-end weakly-supervised single-stage multiple 3D hand mesh reconstruction from a single RGB image.
\newblock \emph{CVIU}, 103706.

\bibitem[{Ren et~al.(2023)Ren, Chen, Hao, Sun, Qi, Wang, and Liao}]{ren2023two}
Ren, P.; Chen, Y.; Hao, J.; Sun, H.; Qi, Q.; Wang, J.; and Liao, J. 2023.
\newblock Two Heads Are Better than One: Image-Point Cloud Network for Depth-Based 3D Hand Pose Estimation.
\newblock In \emph{AAAI}, 2163--2171.

\bibitem[{Romero, Tzionas, and Black(2017)}]{bib:MANO}
Romero, J.; Tzionas, D.; and Black, M.~J. 2017.
\newblock Embodied hands: modeling and capturing hands and bodies together.
\newblock \emph{ACM TOG}, 245:1--245:17.

\bibitem[{Saito et~al.(2019)Saito, Huang, Natsume, Morishima, Kanazawa, and Li}]{saito2019pifu}
Saito, S.; Huang, Z.; Natsume, R.; Morishima, S.; Kanazawa, A.; and Li, H. 2019.
\newblock PIFu: Pixel-Aligned Implicit Function for High-Resolution Clothed Human Digitization.
\newblock In \emph{ICCV}, 2304--2314.

\bibitem[{Walker et~al.(2023)Walker, Mariotti, Vaxman, and Bilen}]{walker2023explicit}
Walker, T.; Mariotti, O.; Vaxman, A.; and Bilen, H. 2023.
\newblock Explicit Neural Surfaces: Learning Continuous Geometry With Deformation Fields.
\newblock \emph{arXiv preprint arXiv:2306.02956}.

\bibitem[{Wang et~al.(2021)Wang, Liu, Liu, Theobalt, Komura, and Wang}]{wang2021neus}
Wang, P.; Liu, L.; Liu, Y.; Theobalt, C.; Komura, T.; and Wang, W. 2021.
\newblock NeuS: Learning Neural Implicit Surfaces by Volume Rendering for Multi-view Reconstruction.
\newblock \emph{NeurIPS}, 34: 27171--27183.

\bibitem[{Weng et~al.(2022)Weng, Curless, Srinivasan, Barron, and Kemelmacher-Shlizerman}]{bib:HumanNeRF}
Weng, C.-Y.; Curless, B.; Srinivasan, P.~P.; Barron, J.~T.; and Kemelmacher-Shlizerman, I. 2022.
\newblock Humannerf: Free-viewpoint rendering of moving people from monocular video.
\newblock In \emph{CVPR}, 16210--16220.

\bibitem[{Xiu et~al.(2022)Xiu, Yang, Tzionas, and Black}]{xiu2022icon}
Xiu, Y.; Yang, J.; Tzionas, D.; and Black, M.~J. 2022.
\newblock {ICON}: {I}mplicit {C}lothed humans {O}btained from {N}ormals.
\newblock In \emph{CVPR}, 13286--13296.

\bibitem[{Zhang et~al.(2021)Zhang, Wang, Deng, Zhang, Tan, Ma, and Wang}]{9710320}
Zhang, B.; Wang, Y.; Deng, X.; Zhang, Y.; Tan, P.; Ma, C.; and Wang, H. 2021.
\newblock Interacting Two-hand 3D Pose and Shape Reconstruction from Single Color Image.
\newblock In \emph{ICCV}, 11354--11363.

\bibitem[{Zuffi et~al.(2017)Zuffi, Kanazawa, Jacobs, and Black}]{bib:smal}
Zuffi, S.; Kanazawa, A.; Jacobs, D.~W.; and Black, M.~J. 2017.
\newblock {3D} Menagerie: Modeling the {3D} Shape and Pose of Animals.
\newblock In \emph{CVPR}, 6365--6373.

\end{thebibliography}

\end{document}